\documentclass[english]{llncs}
\usepackage{color}
\usepackage{array}
\usepackage{verbatim}
\usepackage{rotating}
\usepackage{url}
\usepackage{float}
\usepackage{multirow}
\usepackage{amsmath}
\usepackage{stackrel}
\usepackage{graphicx}
\usepackage{booktabs}
\usepackage{amssymb}

\makeatletter

\providecommand{\tabularnewline}{\\}
\floatstyle{ruled}
\newfloat{algorithm}{tbp}{loa}
\providecommand{\algorithmname}{Algorithm}
\floatname{algorithm}{\protect\algorithmname}

\everymath{\displaystyle}

\makeatother
\usepackage{amsfonts}
\usepackage{babel}
\begin{document}

\title{DeepTract: A Probabilistic Deep Learning Framework for White Matter Fiber Tractography}

\author{Itay Benou \and Tammy Riklin Raviv\inst{1,2}}
%\index{Benou, Itay}
%\index{Tammy, Riklin Raviv}

\institute{}
\institute{Department of Electrical and Computer Engineering,\\
	Ben-Gurion University of the Negev, Beer-Sheva, Israel\\
	\and
	The Zlotowski Center for Neuroscience,\\
	Ben-Gurion University of the Negev, Beer-Sheva, Israel\\
	}

\maketitle
\setcounter{footnote}{0}

\begin{abstract}
We present \emph{DeepTract}, a deep-learning framework for estimating white matter fibers orientation and streamline tractography. We adopt a data-driven approach for fiber reconstruction from diffusion-weighted images (DWI), which does not assume a specific diffusion model. We use a recurrent neural network for mapping sequences of DWI values into probabilistic fiber orientation distributions. Based on these estimations, our model facilitates both deterministic and probabilistic streamline tractography. We quantitatively evaluate our method using the Tractometer tool, demonstrating competitive performance with state-of-the-art classical and machine learning based tractography algorithms. We further present qualitative results of bundle-specific probabilistic tractography obtained using our method. The code is publicly available at: https://github.com/itaybenou/DeepTract.git.

\end{abstract}

\section{Introduction}
Tractography based on diffusion MRI (dMRI) is an important tool in the study of white matter (WM) structures in the brain, allowing to visualize and analyse complex neural tracts in brain connectivity studies~\cite{bullmore2009complex} and investigation of neurological disorders~\cite{benou2018fiber,benou2019combining,ciccarelli2008diffusion}.
Standard tractography pipelines usually consist of a diffusion modeling stage, in which local fiber orientations are estimated from diffusion weighted images (DWI), followed by a tracking stage in which these orientations are translated into WM streamlines. At the heart of the modeling stage lays the problem of finding the local configuration of WM fibers that gave rise to the measured DWI signal. Since a single brain voxel can contain tens of thousands of differently oriented fibers, accurate reconstruction of fiber orientations is a very challenging task. 

Current tractography algorithms can be roughly divided into deterministic approaches, which provide a single streamline orientation in each voxel~\cite{basser1998fiber,lazar2003white}, probabilistic~\cite{berman2008probabilistic}, or global~\cite{fillard2009novel}.  Nevertheless, all of these methods are based on specific, pre-defined, mathematical models for mapping dMRI signals into fiber orientation estimates. Among others, these models include the diffusion tensor model~\cite{basser1994estimation}, Q-ball imaging~\cite{descoteaux2007regularized} and spherical deconvolution \cite{tournier2007robust}. Despite remarkable progress made, current model-based methods are not without limitations~\cite{jeurissen2017diffusion}. Each such model makes specific assumptions regarding WM tissue properties and the dMRI signal, which may vary substantially depending on the subject and the data acquisition process~\cite{neher2015machine}. Some models also impose specific requirements on the data quality and acquisition protocol, e.g., a large number of gradient directions. Therefore, from the user's point of view, choosing a suitable model may prove to be a non-trivial task that requires a high level of expertise.

Machine learning (ML) and deep learning (DL) techniques have demonstrated remarkable abilities in tackling complex problems in a data-driven rather than model-based manner, in a wide variety of domains.
Recently, such approaches have been applied to the task of WM tractography, aiming to directly learn the mapping between input DWI scans and output WM tractography streamlines. By not assuming a specific diffusion model, data-driven algorithms can reduce the dependence on data acquisition schemes and require less user intervention. Neher et al. \cite{neher2015machine} pioneered this line of work, proposing a supervised ML tractography algorithm based on random forest (RF) classifier. The RF classifier was trained to predict a local fiber orientation from a discrete set of possible directions, based on the surrounding dMRI values. More recently, \cite{poulin2017learn} suggested a DL model for fiber tractography, examining a fully-connected (FC) and a recurrent neural network (RNN) architectures. In contrast to \cite{neher2015machine}, streamline tractography was addressed as a regression problem by predicting continuous tracking directions based on sequences of dMRI values. A similar regression approach was presented in \cite{wegmayr2018data} using a multi-layer perceptron (MLP) network. We note that all of these methods perform \textbf{deterministic tractography}, outputting a single streamline direction in each tracking step.
Other DL works have focused strictly on fiber orientation estimation.
For example \cite{koppers2016direct} presented a deep convolutional neural network for estimating discrete fiber orientation distribution functions (fODFs) from dMRI scans. A variation of this idea was presented in~\cite{koppers2017reconstruction}, predicting spherical harmonics coefficients for continuous fODF estimation. These works, however, do not perform fiber tractography.

In this work we present \emph{DeepTract}, a novel DL framework addressing \textit{both} fiber orientation estimation and streamline tractography from DWI scans. To exploit the sequential nature of tractography data, we address the problem as a sequential classification task by training an RNN model to predict local fiber orientations (i.e., classes) along tractography streamlines. Unlike other DL-based tractography algorithms, our model does not output a single deterministic fiber orientation in each tracking step. Instead, it provides a probabilistic estimation of the local fiber orientation distribution in the form of a discrete probability density function. This enables our model to perform deterministic streamline tracking, as well as \textbf{probabilistic tractography} by randomly sampling directions from the estimated distributions. 
We quantitatively evaluate our method using the Tractometer tool~\cite{cote2013tractometer}, demonstrating improved or competitive results compared to state-of-the-art tractography algorithms. We further present qualitative results of high-quality probabilistic tractograms generated by our method.

%\vspace{-0.2cm}
\section{Methods}
In the following sections we describe the proposed DeepTract framework: (1)~the input model, (2)~how the network learns to predict fiber orientations from DWI data, (3)~how new tractograms are generated from unseen data, and (4)~the implementation details of the neural network's architecture.
%\vspace{-0.1cm}

\subsection{Input Model}
%\vspace{-0.1cm}
The training data consists of two sets: a DWI set~$\mathcal{D}$ and its corresponding whole-brain tractography $\mathbf{T}=\left\{ S_{i}\right\} _{i=1}^{N}$ with~$N$ streamlines. Each streamline is represented by a sequence of equi-distant 3D coordinates, i.e., $S_{i}=\left\{ p_{j}\right\} _{j=1}^{n_i}$. 

%\vspace{0.2cm}
\textit{Pre-Processing}: To handle datasets acquired with different gradient schemes, we first resample the DWI set into~$\mathit{K}$ pre-defined gradient directions evenly distributed on the unit hemisphere, using spherical harmonics (we use ~$K$=100). Each DWI volume is then centred according to its mean and normalized by the $b_{0}$ (non-diffusion weighted) volume. 

%\vspace{0.2cm}
\textit{Sequential Input Model}: To make our model invariant to spatial transformations, we feed it with sequences of DWI values instead of directly using the 3D coordinates of the streamlines. Formally, given a DWI dataset~$\mathcal{D}$ and a streamline $S_{i}$, the input to our model is the series of DWI vectors measured along the streamline, i.e. $\left\{ \mathbf{D}(p_{1}),...,\mathbf{D}(p_{n_i})\right\}$. Each input entry $\mathbf{D}(p_{j})$ is a vector of $K$ DWI values measured at location $p_{j}$, such that a single streamline of length $n_i$ corresponds to an input tensor of size $n_i\times K$. 

%\newpage
%\vspace{-0.1cm}
\subsection{Fiber Orientation Estimation}
%\vspace{-0.1cm}
Aiming to facilitate both deterministic and probabilistic tractography, we require our model to provide a probabilistic estimation of local fiber orientations prior to tracking.
For this purpose, we use a discrete representation of an fODF by sampling the unit sphere at $M$ evenly-distributed points $\mathbf{d}=\left\{ d_{m}\right\} _{m=1}^{M}$, each representing a possible fiber orientation. We, therefore, address the problem as a classification task where each orientation is considered as a separate ``class". An additional ``end-of-fiber" (EoF) class is used for labeling streamline termination points. Given an input streamline, our model predicts a probability density function $P(d)$ of $M$$+$1 class probabilities at each point along the streamline. We note that this formulation poses a tradeoff between higher angular resolution, achieved by increasing the number of classes, and the complexity of the classification problem. We used $M$=724, providing an angular resolution of $\sim$3.5$^{\circ}$.

%\vspace{0.2cm}
\textit{Conditional fODFs}: Standard fODFs represent the \textit{total} orientation distribution function at a voxel location $p_{j}$, independent of other voxels, i.e. fODF$_{p_{j}}(d)$ = $P(d)$. Being a sequence-based model, DeepTract has the advantage of utilizing the ``history" of DWI values along an input streamline $S_{i}=\left\{ p_{1},...,p_{n_i}\right\}$. Accordingly, it yields a \textit{conditional} estimation of the fODF at location $p_{j}\in S_{i}$, i.e. \textit{CfODF}$_{\left(p_{j}\mid S_{i}\right)}(d)=P\left(d\mid\mathbf{D}\left(p_{j}\right), \mathbf{D}\left(p_{j-1}\right),...,\mathbf{D}\left(p_{1}\right)\right)$.
We note that a direct relation between CfODFs and total (standard) fODFs can be obtained. Let $prob\left(S_{i}\right)$ denote the probability of reaching the point $p_{j}$ via path $S_{i}$, out of all streamlines passing through $p_{j}$. Using the total probability theorem, we have:
\begin{equation}
\mathit{fODF}_{p_{j}}(d)=\mathbb{E}_{S_{i}}\left[\mathit{CfODF}_{(p_{j}\mid S_{i})}(d)\right]=\underset{i}{\sum}prob\left(S_{i}\right)\mathit{CfODF}_{(p_{j}\mid S_{i})}(d)
\end{equation}  

\begin{center}
	\begin{figure}[t!]
		\centering{}%
		\hspace*{-1.1cm}
		\begin{tabular}{cc}
			\textbf{(a) Training Process} & \textbf{(b) Tracking Process}\tabularnewline
			\includegraphics[scale=0.3]{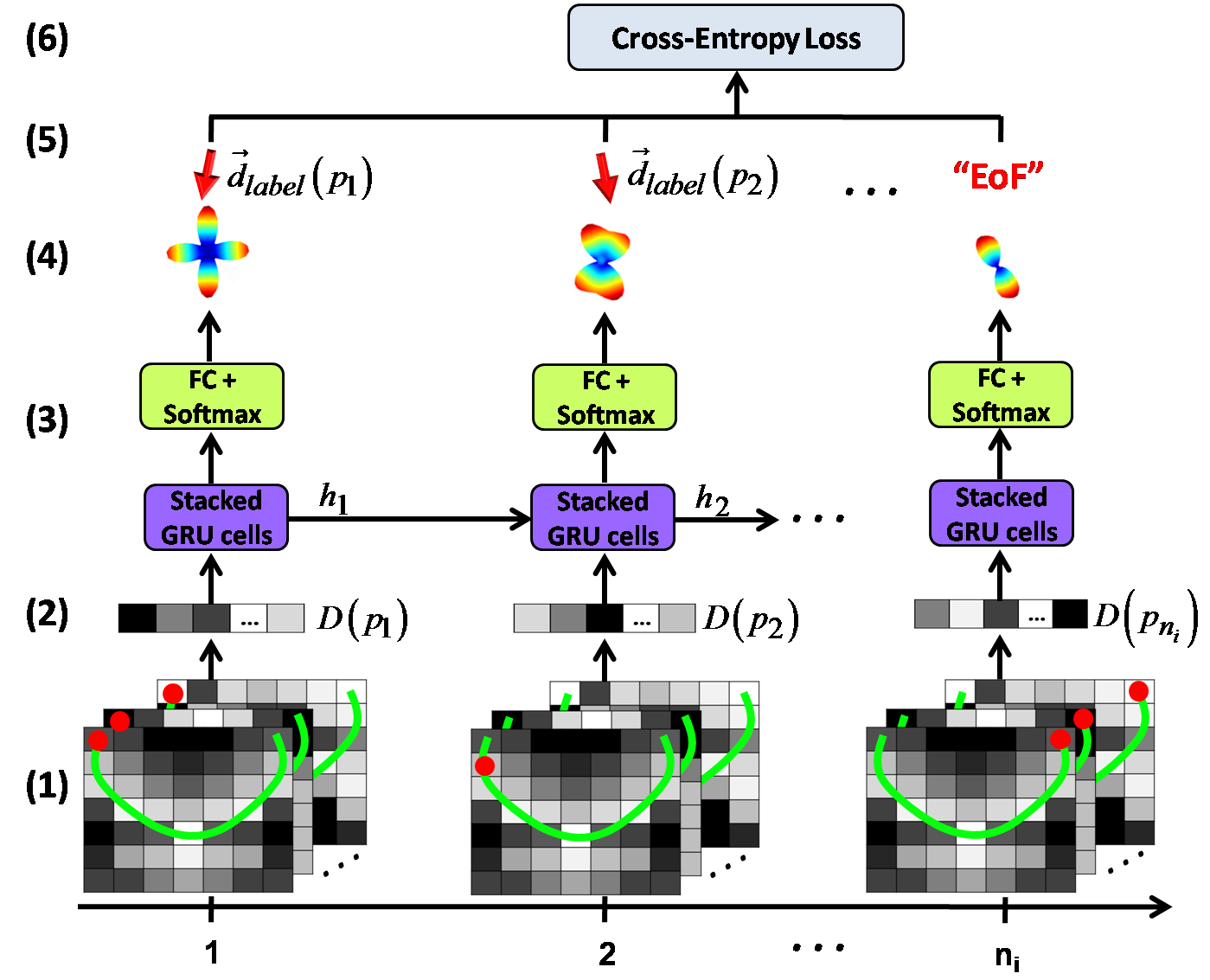} & \includegraphics[scale=0.3]{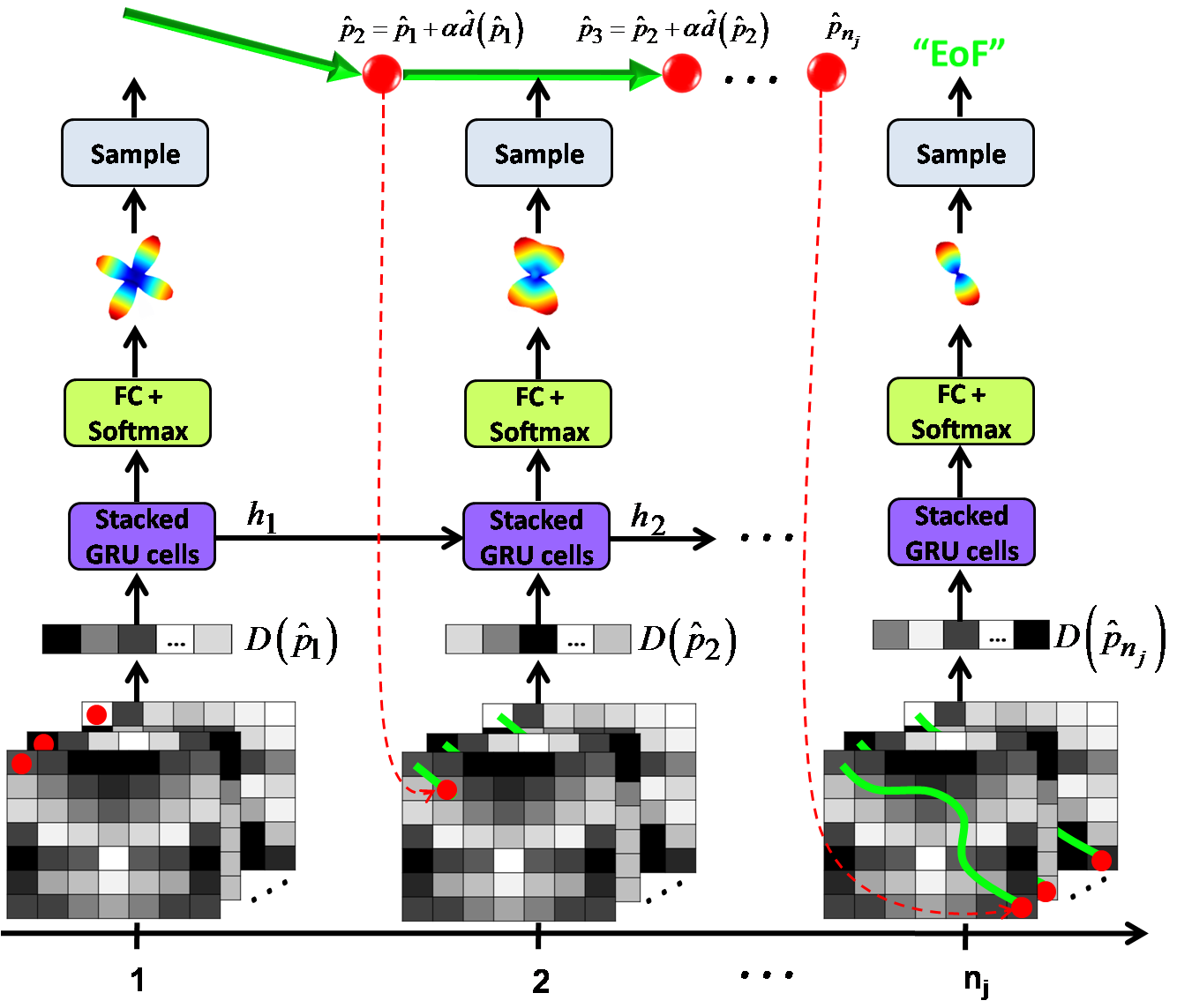}\tabularnewline
		\end{tabular}\caption{Flow of the proposed DeepTract model (unrolled in time) during training (a) and test-time (b). Steps (1)-(4): Sequences of DWI vectors are fed into the network to produce CfODF estimations. (5)-(6): During training (a), predicted CfODFs are compared to the true streamline directions, while at test-time (b) tracking is performed by iteratively sampling from the predicted CfODFs, starting from a given seed point.} 	\label{fig:Network_architecture}
%	\vspace{-1.2cm}
	\end{figure}
	\par\end{center}

\vspace{-0.8cm}
\subsection{Streamline Tractography}
%\vspace{-0.2cm}
During training, the predicted CfODF at location $p_{j}$ is compared to the ``true" orientation defined by $d_{label}(p_{j})=(p_{j+1}-p_{j})/\left\Vert p_{j+1}-p_{j}\right\Vert $ (see Fig.~\ref{fig:Network_architecture}(a)). The corresponding class label $y_{p_{j}}(d)$ is the orientation $d_{m}\in\mathbf{d}$ which is closest to $d_{label}(p_{j})$, represented by a ``1-hot" vector. Once the model is trained, streamline tractography can be performed on unseen DWI scans in an iterative process as illustrated in Fig.~\ref{fig:Network_architecture}(b). Given an initial seed point $\hat{p_{1}}$, the corresponding DWI vector $\mathbf{D}(\hat{p_{1}})$ is fed into the network, which in turn provides CfODF($\hat{p_{1}}$) as output. \textit{Deterministic} tracking is performed by stepping in the most likely fiber orientation, i.e. $\hat{d}(\hat{p_{1}})=\arg\underset{d\in\mathbf{d}}{\max}\mathit{CfODF}(\hat{p_{1}})$. Alternatively, \textit{probabilistic} tracking can be performed by randomly sampling a direction from the CfODF. Either way, the streamline is propagated iteratively according to $\hat{p}_{j+1}=\hat{p}_{j}+\alpha\hat{d}(p_{j})$, where $\alpha$ is the step size. The process is repeated until the EoF class is predicted.

%\vspace{-0.4cm}
\subsection{Network Architecture and Loss Function}
%\vspace{-0.2cm}
We implement our model using an RNN, specifically a Gated Recurrent Unit (GRU) \cite{chung2014empirical}. The proposed network consists of five stacked GRU layers, each containing 1000 units. We use ReLU activations for all layers but the last one, which is a fully-connected (FC) layer followed by a softmax operation. The loss function for a single input streamline $S_{i}$ is the mean cross-entropy between the predicted CfODFs and the true labels along the streamline:
\begin{equation}
\mathcal{L_{\mathit{i}}}=-\frac{1}{n_{i}}\stackrel[j=1]{n_{i}}{\sum}\stackrel[m=1]{M+1}{\sum}y_{p_{j}}\left(d_{m}\right)\mathrm{log}\left(CfODF_{\left(p_{j}\mid S_{i}\right)}\left(d_{m}\right)\right)
\end{equation} 

%\vspace{-0.4cm}
\subsubsection{Label Smoothing:}
\begin{figure}[b]
	\centering{}%
	\begin{tabular}{cccc}
		\textbf{Original Label~} & \textbf{~Smooth Label} &  & \tabularnewline
		\includegraphics[scale=0.5]{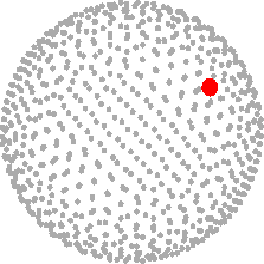}~~ & ~~\includegraphics[scale=0.5]{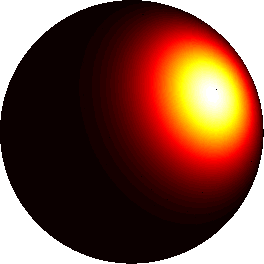} & \includegraphics[scale=0.35]{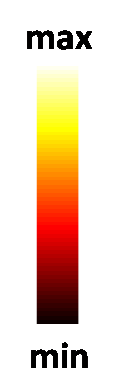} & \begin{turn}{90}
			{\footnotesize{}~~~Probability}
		\end{turn}\tabularnewline
	\end{tabular}\caption{Label smoothing. The original sparse label (left) is smoothed using
		a Gaussian kernel on the unit sphere, transferring probability mass
		to neighboring classes (right). } \label{fig:label_smoothing}
\end{figure}

Unlike traditional classification tasks, here the classes are geometrically structured with a well-defined angular metric. This implies that different classifications errors should be weighted according to this metric, thus a 1-hot label (i.e., a delta function) is not suitable. To account for the spatial structure of the classes, we propose to smooth the true labels by convolving them with a Gaussian kernel $G$ of width $\tau$ on the unit sphere:

\begin{equation}
y_{smooth}\left(d\right)=\delta\left(d-d_{label}\right)*G\left(d\right)=\frac{1}{Z}\exp\left(-\frac{\measuredangle\left(d,d_{label}\right)}{\tau}\right)
\end{equation}
where ${\measuredangle\left(d,d_{label}\right)}$ is the angle between a direction $d$ and the ground truth direction $d_{label}$, and $Z$ is a normalization constant (see Fig. \ref{fig:label_smoothing}).

\vspace{-0.4cm}\subsubsection{Entropy-Based Tracking Termination:}
During the generative process of RNNs at test time, accumulated error may divert predictions from the training data distribution \cite{graves2013generating}, towards ``unfamiliar" input values.
This may result in an increased uncertainty of our model's predictions, manifesting as more isotropic CfODF estimations and erroneous tracking steps. 
To alleviate this problem, we introduce an entropy-based tracking termination criterion, in addition to the EoF class. Specifically, we terminate the tracking process whenever the entropy of a predicted CfODF exceeds the following dynamic threshold~$E_{\mbox\scriptsize{th}}\left(t\right)=a\exp\left(-t/b\right)+c$, where $t$ is the sequence time step, and $a$,$b$ and $c$ are hyperparameters.

%\vspace{-0.2cm}
\section{Experiments and Results}
%\vspace{-0.1cm}
To test the performance of the proposed method we used the ISMRM tractography challenge DWI phantom dataset \cite{maier2016tractography}. Our experiments include: 1)~Quantitative evaluation of whole brain tractography using the Tractometer tool.
2)~Qualitative (visual) demonstration of bundle-specific probabilistic tractography performed by our model. 

\textit{Pre-Processing:}
DWIs were denoised \cite{manjon2013diffusion} and corrected for eddy currents and head motion. For supervision, whole brain tractography was performed using Q-ball reconstruction \cite{aganj2009odf} followed by probabilistic tracking using the MITK diffusion tool. The resulting streamlines were divided into training and validation sets using a 90\%-10\% split. Data augmentation was performed by reversing the orientation of all streamlines in the training set, resulting in $\sim$400K streamlines.

\textit{Training Procedure:}
Training was performed using the Adam optimizer with a batch size of 32 streamlines. To avoid overfitting, dropout was used with deletion probability of 0.3, as well as gradient clipping to avoid exploding gradients.

%\vspace{-0.1cm}
\subsection{Whole Brain Tractography Evaluation}
%\vspace{-0.1cm}
We first evaluate our method using Tractometer - a publicly available online tool for assessment of whole brain tractography algorithms.

\textit{Tractography:} We used our trained model to perform whole brain deterministic tractography. Tracking was initialized by randomly placing 200K seed points within the DWI volume. Using the validation set, the step-size was set to $\alpha=0.5$ (in voxels), and the entropy-based stopping parameters were set to $a$=3, $b$=10 and $c$=4.5. Tracking was terminated for high-curvature steps (over 60$^{\circ}$), and output streamlines longer than 200mm and shorter than 20mm were discarded.

\textit{Evaluation:} Tractometer evaluates a whole brain tractogram by comparing it to 25 gold standard streamline bundles, outputting the following scores: three measures for the correctness of streamline connectivity, i.e. percentage of valid connections (VC), invalid connections (VC) and non-connections (NC); two measures for the correctness of identified bundles, i.e. number of valid and invalid bundles (VB and IB); and three scores for the correctness of voxel coverage, i.e. overlap (OL), overreach (OR) and $F_1$ scores. Please refer to \cite{cote2013tractometer} for more details.

\textit{Results:} The whole brain tractography generated by the proposed DeepTract method is shown in Fig. \ref{fig:WB_tractographies}, along with those of MITK and the ISMRM challenge gold standard. The Tractometer scores of our method, when supervised by MITK and by the gold-standard tractography, are summarized in Table~\ref{Table: Tractometer}. We compare these results
to the performance of MITK (supervisor), the average ISMRM challenge submission, and two other DL tractography methods \cite{poulin2017learn,wegmayr2018data}. 
The results demonstrate that DeepTract is competitive or outperforms the examined methods in most parameters.
Specifically, when using the gold standard tractography for supervision, DeepTract demonstrated the best voxel coverage performance, i.e. highest OL and F1 scores. In addition, DeepTract scored the best false-positive connections rates (lowest IC and NC), and was the only method to successfully detect all 25 bundles – clearly demonstrating the high-limit capability of the proposed method.
We note that our VC score is slightly lower than \cite{wegmayr2018data}, however \cite{wegmayr2018data} did not report their IC, NC and F1 score, so a complete comparison to their work is not possible.
We further note that even when using the MITK tractography for supervision, our method demonstrated good overall performance, including the best false-positive rates (lowest OR and IB). This result is probably due to our entropy-based termination criterion, which prevents streamlines from straying off coherent bundle structures.\vspace{-0.5cm}

\begin{center}
	\begin{figure}[t!]
		\begin{centering}
			\begin{tabular}{ccc}
				\textbf{DeepTract (proposed)} & \textbf{MITK} & \textbf{Ground Truth}\tabularnewline
				\includegraphics[scale=0.13]{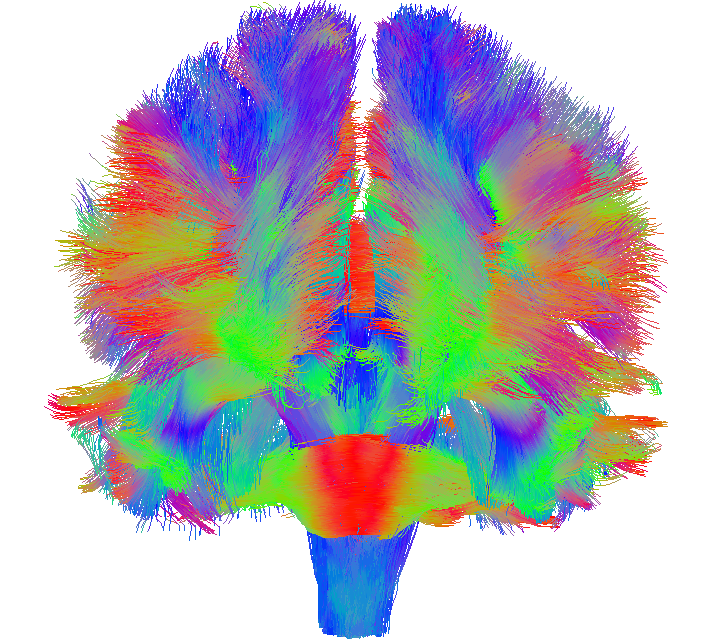} & \includegraphics[scale=0.13]{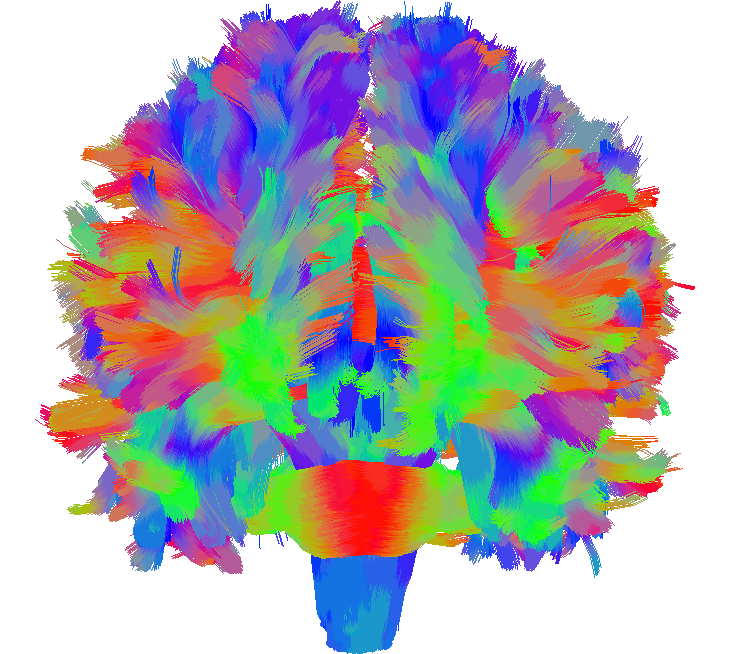} & \includegraphics[scale=0.13]{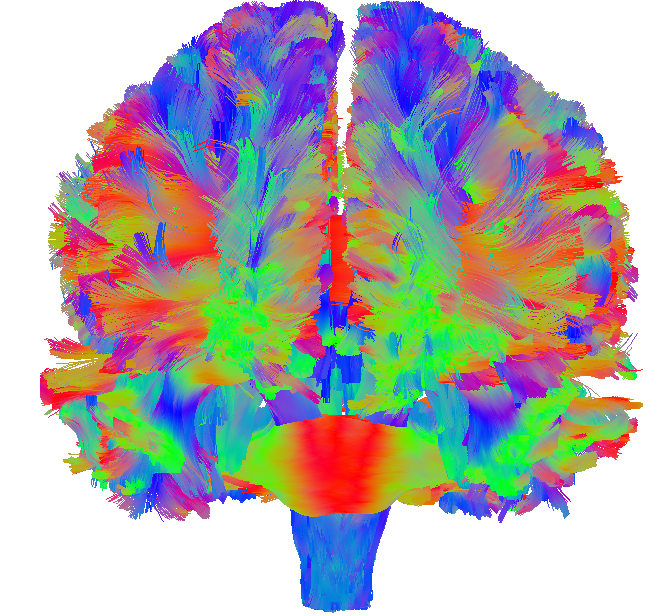}\tabularnewline
				\includegraphics[scale=0.115]{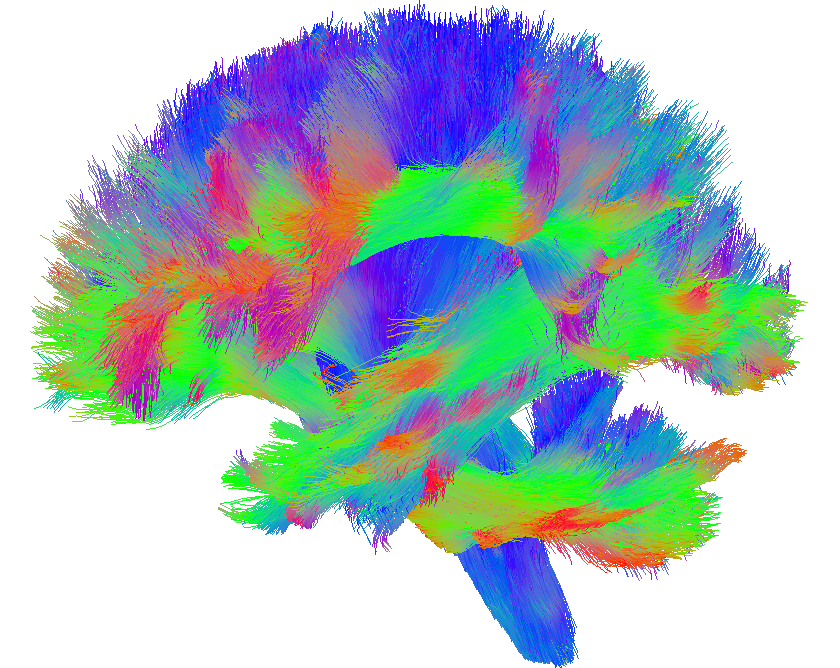} & \includegraphics[scale=0.115]{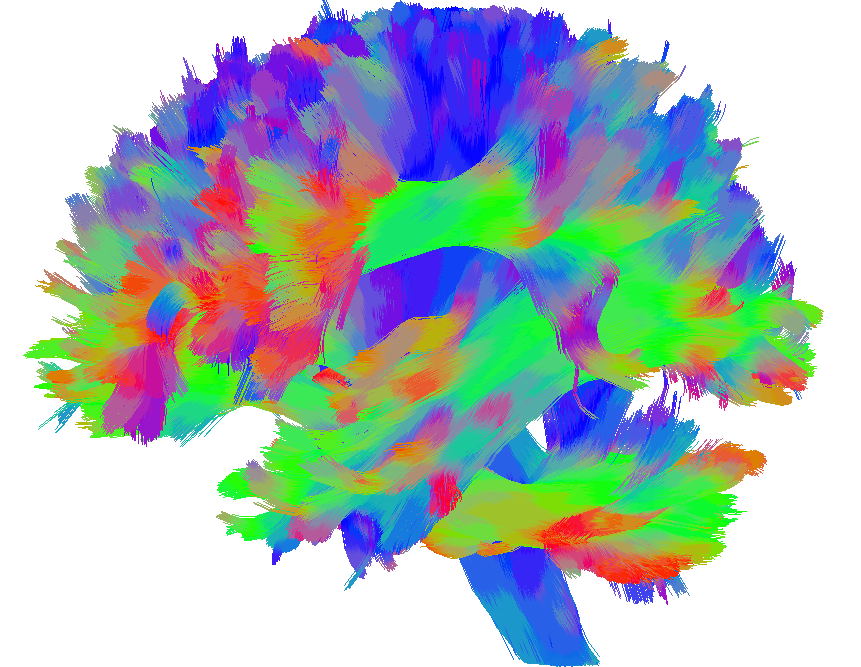} & \includegraphics[scale=0.115]{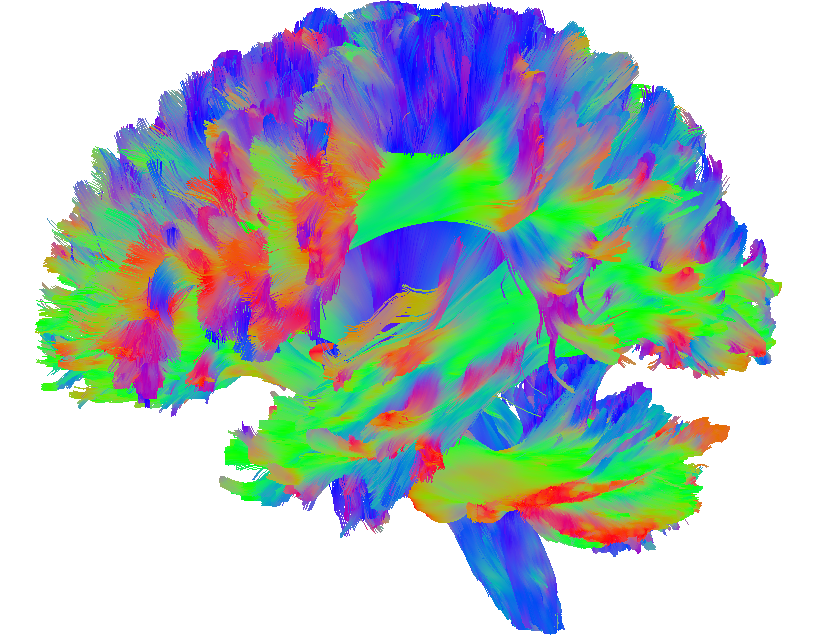}\tabularnewline
			\end{tabular}\caption{Whole brain tractography results - visual comparison.} \label{fig:WB_tractographies}
%		\vspace{-0.4cm}
			\par\end{centering}
%		\vspace{-0.1cm}
	\end{figure}
	\par\end{center}

	\begin{table}[t!]
		\begin{centering}
			\begin{tabular}{l|cccccccc}
				\hline 
				\multirow{2}{*}{\textbf{Model}} & \multicolumn{3}{c}{Connections (\%)} & \multicolumn{2}{c}{Bundles} & \multicolumn{3}{c}{Coverage (\%)}\tabularnewline
				& \textbf{~VC$\uparrow$~} & \textbf{~IC$\downarrow$~} & \textbf{~NC$\downarrow$~} & \textbf{~VB$\uparrow$~} & \textbf{~IB$\downarrow$~} & \textbf{~OL$\uparrow$~} & \textbf{~OR$\downarrow$~} & \textbf{~~F1$\uparrow$}\tabularnewline
				\hline 
				ISMRM mean results & 53.6  & 19.7 & 25.2 &  21.4 & 281 & 31.0 & 23.0 & 44.2\tabularnewline
				\hline 
				Poulin et al. \cite{poulin2017learn} & 41.6 & 45.6 & 12.8 & 23 & 130 & 64.4 & 35.4 & 64.5\tabularnewline
				\hline 
				Wegmayr et al. \cite{wegmayr2018data} & \textbf{72} & - & - & 23 & 57 & 16.0 & 28.0 & -\tabularnewline
				\hline 
				MITK (supervisor) & 59.1  & 27.8 & 13.1 & 24 & 69 & 47.2 & 31.2 & 52.5\tabularnewline
				\hline 
				\hline 
				\textbf{Proposed} (GT supervision) & 70.6 & \textbf{19.5} & \textbf{9.9} & \textbf{25} & 56 & \textbf{69.3} & 22.7 & \textbf{70.1}\tabularnewline
				\hline 
				\textbf{Proposed} (MITK supervision) & 40.5 & 32.6 & 22.9 & 23 & \textbf{51} & 34.4 & \textbf{17.3} & 44.2\tabularnewline
				\hline 
			\end{tabular}
			\par\end{centering}
		\vspace{0.1cm}
		\caption{Tractometer evaluation results. Up-arrow represents higher-is-better metrics, while down arrow represents lower-is-better ones. Best scores are in bold font.} \label{Table: Tractometer}
		\vspace{-0.4cm}
	\end{table} 

%\vspace{-0.7cm}
\subsection{Probabilistic Tracking}
\begin{figure}[t!]
	\centering{}%
	\hspace*{-0.22cm}
	\begin{tabular}{ccccc}
		\multicolumn{2}{c}{Frontopontine Tract} &  & \multicolumn{2}{c}{Uncinate Fasciculus}\tabularnewline
		\textbf{Ground Truth} & \textbf{Proposed} & \textbf{Prob.} & \textbf{Proposed} & \textbf{Ground Truth}\tabularnewline
		\includegraphics[scale=0.3]{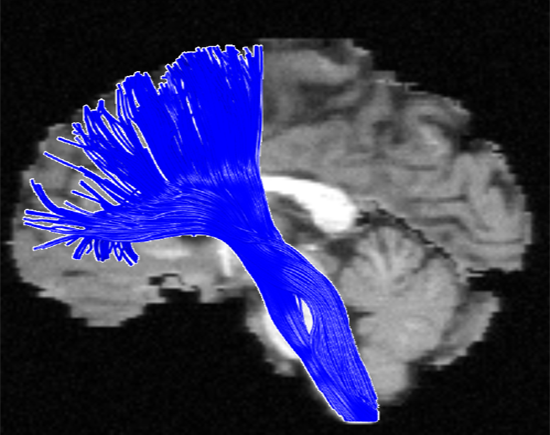} & \includegraphics[scale=0.3]{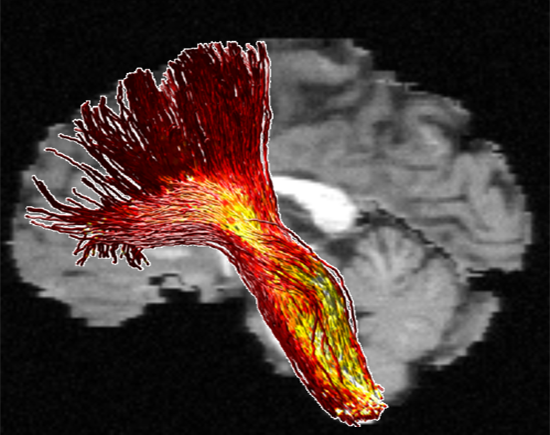} & \includegraphics[scale=0.35]{figures/labels_colorbar2.png} & \includegraphics[scale=0.3]{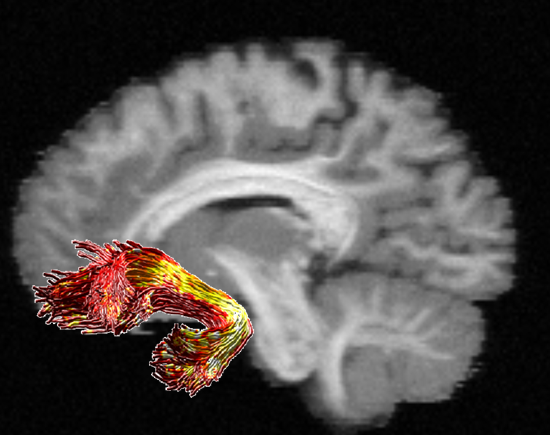} & \includegraphics[scale=0.3]{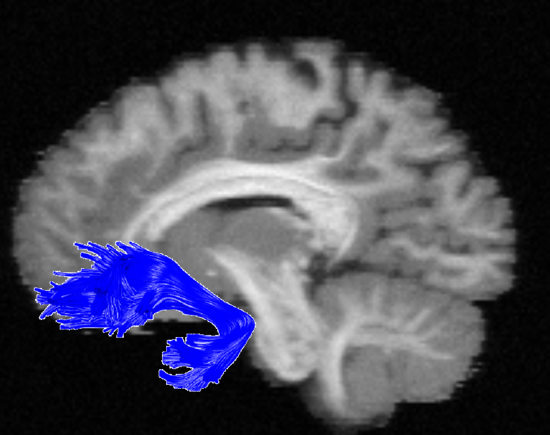}\tabularnewline
	\end{tabular}\caption{Bundle-specific probabilistic tractography results of the proposed method.} \label{fig:prob_tractographies}
%\vspace{-0.1cm}
\end{figure}

%\vspace{-0.1cm}
We further demonstrate DeepTract's ability to perform probabilistic tractography, using the phantom DWI dataset of the ISMRM challenge. Bundle-specific tracking was performed by seeding from the endpoints of a gold-standard bundle, and then sampling iteratively from the predicted CfODFs. The process was repeated $T=20$ times to create a probabilistic map counting the number of ``visits" in each voxel. Results for the Frontopontine tract and the Uncinate Fasciculus are shown in Fig. \ref{fig:prob_tractographies}, alongside the ground truth bundles. Visual evaluation shows that the resulting bundles are in-line with the ground truth tractograms.
Also note that higher probabilities were assigned to the bundles' core, gradually decreasing as fibers diverge towards their endpoints due to higher uncertainty.

%\vspace{-0.1cm}
\section{Summary and Discussion}
%\vspace{-0.1cm}
We presented DeepTract, the first deep learning method capable of performing \textit{both deterministic and probabilistic} tractography from DWI data. We showed that by combining an RNN-based sequential approach with a discrete classification framework, our model provides reliable probabilistic fiber orientation estimations. In a quantitative evaluation, the proposed method outperformed or was competitive with state-of-the-art classical and DL tractography algorithms.
While larger dMRI and tractography datasets are needed to further progress the research of data-driven tractography, the results obtained in this work demonstrate the potential of DL methods for WM tractography applications.

%\vspace{-0.2cm}
\bibliographystyle{plain}
\bibliography{RefTractography}

\end{document}